\documentclass[lettersize,journal]{IEEEtran}
\usepackage{amsmath,amsfonts}
\usepackage{algorithmic}
\usepackage{array}
\usepackage[caption=false,font=normalsize,labelfont=sf,textfont=sf]{subfig}
\usepackage{textcomp}
\usepackage{stfloats}
\usepackage{url}
\usepackage{verbatim}
\usepackage{graphicx}
\usepackage{xcolor}

\usepackage{graphics} 
\usepackage{epsfig} 
\usepackage{mathptmx} 
\usepackage{times} 
\usepackage{amsmath,etoolbox}
\AtBeginEnvironment{pmatrix}{\setlength{\arraycolsep}{2pt}}
\usepackage{amssymb}  
\usepackage{eucal}
\usepackage{enumitem,kantlipsum}

\def\BibTeX{{\rm B\kern-.05em{\sc i\kern-.025em b}\kern-.08em
T\kern-.1667em\lower.7ex\hbox{E}\kern-.125emX}}
\usepackage{balance}

\begin{document}

\title{Pitch Angle Control of a Magnetically Actuated Capsule Robot with Nonlinear FEA-based MPC and EKF Multisensory Fusion}
\author{Chongxun Wang$^{1}$, Zikang Shen$^{1}$, Apoorav Rathore$^{1}$, Akanimoh Udombeh$^{1}$, Harrison Teng$^{1}$ and Fangzhou Xia$^{1*}$
\vspace{-3em}
\thanks{$^{1}$ The University of Texas at Austin, Walker Department of Mechanical Engineering, MINIMAX Lab, 78712, Austin, TX, USA}%
\thanks{$^{*}$ Corresponding author: Fangzhou Xia, address: 204 E  Dean Keeton St, ETC 4.158, AUSTIN, TX, 78712; phone: +1(512)232-2821; email: {\tt\small fangzhou.xia@austin.utexas.edu}}
\thanks{${\dag}$ These authors contributed equally to this work.} 
}
\markboth{2026 IEEE/ASME Transactions on Mechatronics}
{How to Use the IEEEtran \LaTeX \ Templates}
\maketitle

\begin{abstract}
Magnetically actuated capsule robots promise minimally invasive diagnosis and therapy in the gastrointestinal (GI) tract, but existing systems largely neglect control of capsule pitch, a degree of freedom critical for contact-rich interaction with inclined gastric walls. This paper presents a nonlinear, model-based framework for magnetic pitch control of an ingestible capsule robot actuated by a four-coil electromagnetic array. Angle-dependent magnetic forces and torques acting on embedded permanent magnets are characterized using three-dimensional finite-element simulations and embedded as lookup tables in a control-oriented rigid-body pitching model with rolling contact and actuator dynamics. A constrained model predictive controller (MPC) is designed to regulate pitch while respecting hardware-imposed current and slew-rate limits. Experiments on a compliant stomach-inspired surface demonstrate robust pitch reorientation from both horizontal and upright configurations, achieving about three to five times faster settling and reduced oscillatory motion than on–off control. Furthermore, an extended Kalman filter (EKF) fusing inertial sensing with intermittent visual measurements enables stable closed-loop control when the camera update rate is reduced from 30 Hz to 1 Hz, emulating clinically realistic imaging constraints. These results establish finite-element-informed MPC with sensor fusion as a scalable strategy for pitch regulation, controlled docking, and future multi-degree-of-freedom capsule locomotion.
\end{abstract}
\vspace{-2mm}


\vspace{-6mm}
\section{Introduction}
\vspace{-1mm}
Ingestible capsule robots are an increasingly important class of minimally invasive medical devices for gastrointestinal (GI) diagnosis and therapy, enabling applications such as capsule endoscopy, motility assessment, targeted drug delivery, and localized sampling \cite{iddan2000wireless, singeap2016capsule, cao2024capsule}. Compared to conventional tethered endoscopy, capsule-based systems promise reduced patient discomfort and improved accessibility, motivating extensive research into active locomotion and control strategies \cite{abbott2020magnetic, kim2022magneticsoft}. However, most clinically deployed capsules remain largely passive, relying on peristalsis and gravity for transport, which limits their ability to reach specific anatomical sites or maintain stable contact with regions of interest.

Magnetic actuation has therefore emerged as a dominant approach for enabling wireless control of ingestible robots, owing to its deep tissue penetration and compatibility with clinically acceptable field strengths \cite{ceylan2017mobile, abbott2020magnetic}. By embedding permanent magnets within the capsule and generating external magnetic fields using coils or robotic magnets, prior studies have demonstrated a range of motion primitives, including rolling, steering, and helical swimming \cite{peyer2013bioinspired, li2018development}. Nevertheless, most existing magnetic capsule systems primarily address \emph{in-plane} motion control—roll and yaw—on flat or gently inclined surfaces \cite{yim2011design, son2020magnetically}, including our recent work on magnetically driven capsule locomotion \cite{Zhou2026Anisotropic}.

In the stomach, navigation is fundamentally more demanding than in conventional capsule endoscopy. While existing capsule endoscopes rely on brief, non-contact reorientation to visually scan the lumen, emerging therapeutic and diagnostic tasks—such as localized drug delivery or targeted injection—require sustained, stable contact with the gastric wall. The stomach’s steep local slopes, mucosal folds, and irregular three-dimensional geometry render in-plane control alone insufficient. In this setting, active regulation of capsule \emph{pitch} becomes essential for aligning with inclined or near-vertical surfaces, enabling contact-rich interaction, and traversing uneven terrain beyond the capabilities of roll and yaw alone. Furthermore, combining the pitch dynamic model and on-board inertia sensors can provide information for state estimation using sensor-fusion, allowing reduced reliance on high-rate external imaging—an important consideration for ingestible systems where optical sensing may be impractical and imaging rates for X-ray should be minimized.

Fig.~\ref{fig:system_overview} illustrates the problem setting considered in this work. A capsule robot containing internal permanent magnets is actuated by four external electromagnetic coils while moving on a compliant, silicone-based surface that approximates a local patch of the gastric wall. By actively regulating pitch, the capsule can dock against inclined surfaces and maneuver across localized protruding topographic features that are difficult for pure rolling locomotion. Building on this setting, we investigate the modeling, control, and sensing aspects of magnetic pitch regulation for an ingestible capsule robot operating on stomach-inspired surfaces. Finite-element magnetostatic simulations are used to characterize the nonlinear, angle-dependent magnetic forces and torques acting on the embedded magnets, which are condensed into torque-per-ampere lookup tables and embedded in a control-oriented rigid-body pitch dynamic model incorporating gravity, rolling contact kinematics, and current-driver dynamics. A constrained model predictive controller (MPC) is then designed to exploit this nonlinear actuation model while enforcing current amplitude and slew-rate limits. The proposed framework is evaluated through controlled pitch reorientation experiments from different initial configurations, and through sensing studies that compare camera-only feedback against EKF-based fusion of onboard inertial sensing with intermittent visual updates, highlighting the trade-offs between sensing rate, estimation accuracy, and closed-loop control performance under realistic constraints. The main contributions of this work are threefold:

\begin{figure}[t]
  \centering
  \includegraphics[width=\linewidth]{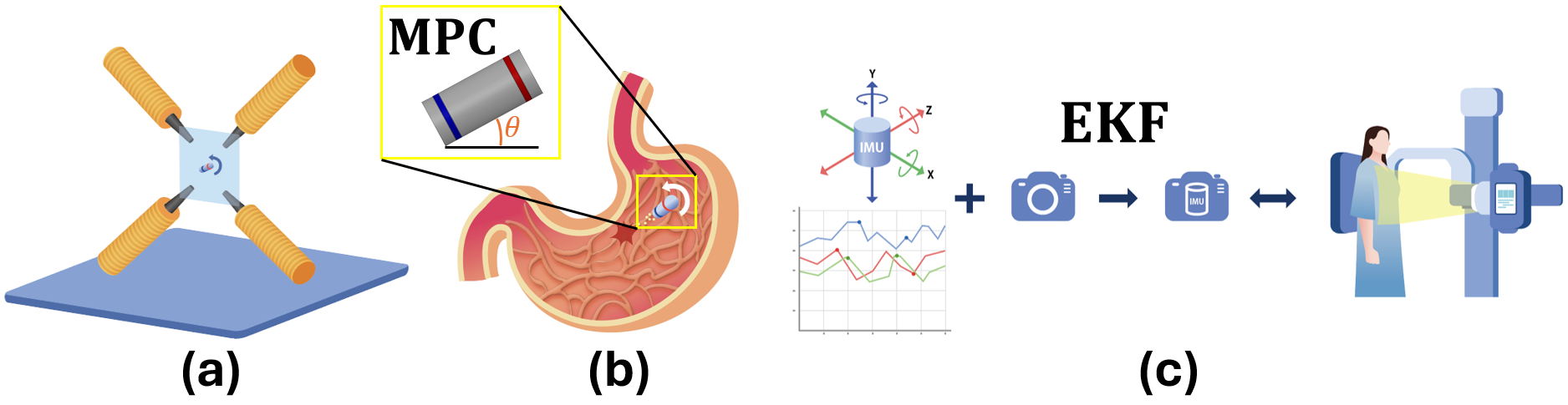}
  \vspace{-9mm}
  \caption{System overview of magnetically actuated pitch control for an ingestible capsule robot. (a) Four external electromagnetic coils generate controlled magnetic fields that interact with internal permanent magnets. (b) The magnetic field produce torques that regulate capsule pitch as it rolls on a soft stomach phantom to enable docking on inclined gastric walls. (c) State estimation through fusion of onboard inertial sensing and downsampled camera-based angle measurements, which emulates low-rate X-ray.}
  \label{fig:system_overview}
  \vspace{-5mm}
\end{figure}

\begin{itemize}
    \item[(i)] The design and experimental validation of a magnetically actuated capsule platform tailored for contact-rich pitch control on gastric phantom, demonstrating robust pitch reorientation in different initial configurations;
    \item[(ii)] A control-oriented nonlinear magneto--mechanical pitch dynamics model that integrates finite-element force and torque lookup table with rolling-contact kinematics, and a constrained MPC  that achieves approximately three to five times faster settling and substantially reduced oscillatory motion compared to on--off control;
    \item[(iii)] An integrated sensing and control architecture that enables stable pitch regulation while reducing the effective camera measurement rate from 30~Hz to 1~Hz through EKF-based fusion of intermittent visual measurements and onboard inertial sensing.
\end{itemize}

The remainder of this paper is organized as follows. Section~\ref{sec:system} describes the capsule design, magnetic actuation hardware, and sensing architecture. Section~\ref{sec:model_control} presents the finite-element-based magneto--mechanical modeling framework, the MPC formulation, and the state estimation approach. Section~\ref{sec:experiments} reports experimental validation results under multiple control and sensing configurations. Finally, Section~\ref{sec:conclusion} concludes with a discussion of implications and future directions.

\vspace{-4mm}
\section{Mechatronic System Design}
\label{sec:system}
\vspace{-1mm}
This section describes the integrated mechatronic system developed for magnetic pitch angle control of an ingestible capsule robot on stomach-inspired surfaces. Experiments are conducted on a 3D-printed silicone surface with Shore~A hardness of 18, chosen to approximate the compliance of the gastric interior and to reproduce contact conditions relevant to docking and reorientation. The overall hardware geometry is summarized in Fig.~\ref{fig:hardware_geometry} and the embedded electronics are shown in Fig.~\ref{fig:embedded_electronics}. Finite-element magnetostatic simulations are used to characterize angle-dependent magnetic forces and torques for control modeling (Section~\ref{subsec:fea_map}). Finally, the sensing and real-time control architecture is summarized in Fig.~\ref{fig:realtime_arch}.

\vspace{-4mm}
\subsection{Capsule Mechanical Design}
\label{subsec:capsule_mech}
\vspace{-1mm}

\begin{figure}[t]
  \centering
  \includegraphics[width=\linewidth]{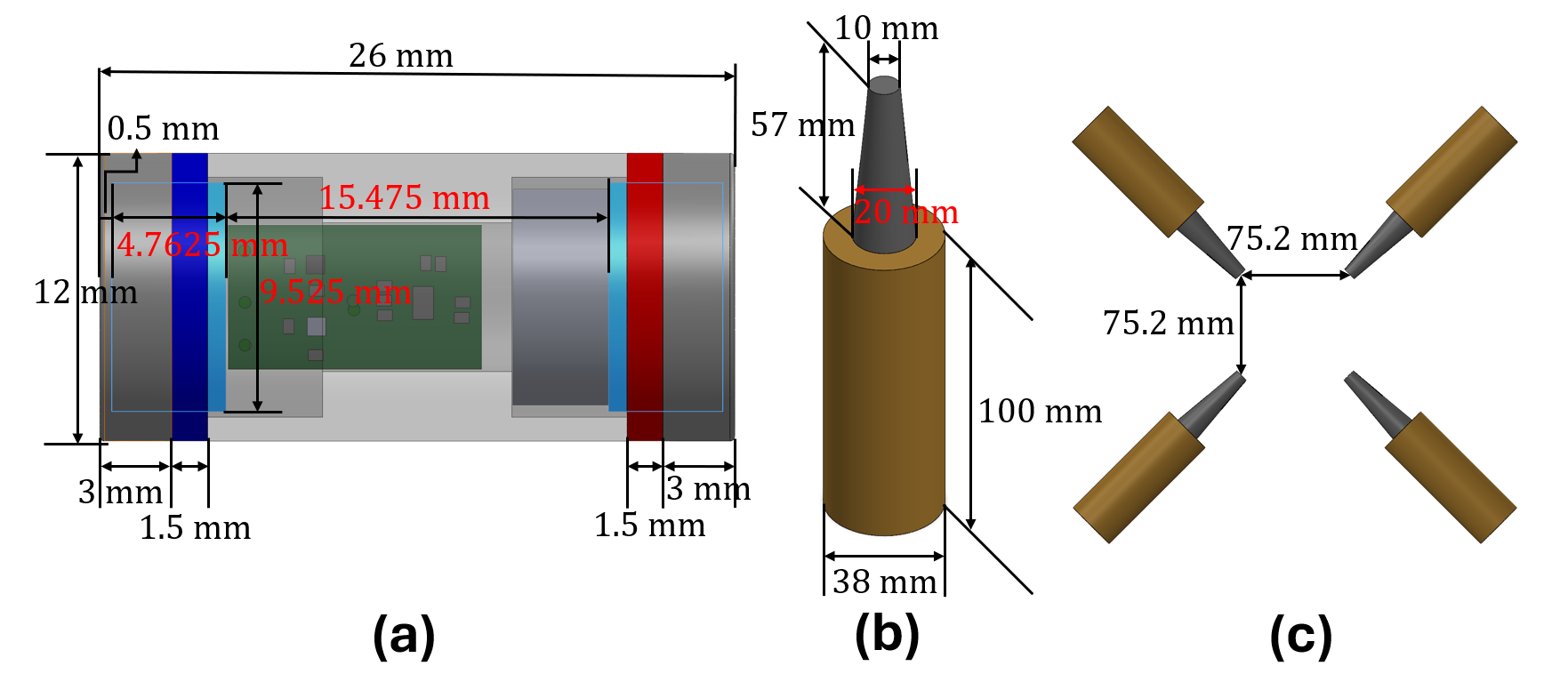}
  \vspace{-9mm}
  \caption{Combined hardware geometry of the capsule robot and electromagnetic coil assembly. (a): Capsule dimensions and internal magnet placement, including colored fiducial rings used for vision-based pitch estimation. (b): Geometry of a single electromagnetic coil with conical iron core. (c) Front-view of the four-coil arrangement defining the operating workspace.}
  \label{fig:hardware_geometry}
  \vspace{-5mm}
\end{figure}

As shown in Fig.~\ref{fig:hardware_geometry}(a), the capsule body is a cylindrical shell of total length $26~\mathrm{mm}$ and outer diameter $12~\mathrm{mm}$, fabricated via 3D printing. These dimensions are chosen to be consistent with those of commercially available capsule endoscopes, ensuring that the proposed mechanical design remains representative of clinically relevant size constraints.

A pair of identical neodymium permanent magnets (3/8~inch diameter $\times$ 3/16~inch length, grade N52) are mounted coaxially inside the capsule and magnetized along the capsule axis with the same polarity. Each magnet is positioned with a $0.5~\mathrm{mm}$ axial clearance from the corresponding capsule end. The two magnets are separated by an axial distance of $15.475~\mathrm{mm}$ between their inner faces. This symmetric placement ensures balanced mass distribution and cooperative magnetic torque generation during pitch actuation.

The space between the two magnets houses the embedded electronics stack, consisting of two printed circuit boards (PCBs), each measuring $6~\mathrm{mm} \times 10~\mathrm{mm} \times$ minimal PCB thickness, and a $3.7~\mathrm{V}$ lithium battery with $9~\mathrm{mm}$ diameter and $4~\mathrm{mm}$ thickness. The electronics and battery are arranged coaxially within the capsule, with the IMU sensor, as described in Section~\ref{subsec:electronics}, positioned at the geometric center of the capsule to accurately measure its motion, while minimizing perturbations to the rolling and pitching dynamics.

For vision-based pitch estimation, two high-contrast fiducial bands are applied to the capsule exterior: a blue ring located $3~\mathrm{mm}$ from the proximal end and a red ring located $3~\mathrm{mm}$ from the distal end, each with an axial width of $1.5~\mathrm{mm}$. A front-facing camera tracks these fiducials to estimate the capsule pitch angle during experiments, as described in Section~\ref{subsec:integration_realtime}.

\vspace{-4mm}
\subsection{Magnetic Coil Actuation System Design}
\label{subsec:coil_design}
\vspace{-1mm}

Pitch actuation is provided by four identical electromagnetic coil--core assemblies arranged around the operating surface (Fig.~\ref{fig:hardware_geometry}(b,c)). Each coil consists of 1500 turns of 0.7 mm diameter copper wire wound on a cylindrical bobbin (inner diameter $20~\mathrm{mm}$, outer diameter $38~\mathrm{mm}$, axial length $100~\mathrm{mm}$) with a soft iron core that extends toward the workspace as a conical pole piece (base diameter $20~\mathrm{mm}$, tip diameter $10~\mathrm{mm}$, height $57~\mathrm{mm}$). The four cone tips are oriented toward the workspace center and arranged along the diagonals of a square, with a center-to-center spacing of $75.2~\mathrm{mm}$ in both orthogonal directions, defining a $75.2~\mathrm{mm} \times 75.2~\mathrm{mm}$ operating region.

The coil geometry, number of turns, and inter-coil spacing were determined through finite-element simulations in ANSYS Maxwell, which evaluated achievable magnetic field strength and torque distribution under a current constraint of approximately $1~\mathrm{A}$ imposed by the limit of the current driver. The resulting design balances usable actuation workspace with sufficient magnetic torque for reliable pitch reorientation.

The actuation strategy emphasizes control of the magnetic field direction for attitude regulation rather than strong field gradients for translation. In pitch-control experiments, energizing a diagonal pair of coils generates an approximately diagonal field that produces pitching torques on the internal permanent magnets. Driving all four coils symmetrically produces a predominantly vertical field, which is used to re-initialize the capsule near upright configurations and provides uniform-field alignment effects (e.g., Helmholtz-like alignment).

\vspace{-5mm}
\subsection{Embedded Electronics and Inertial Sensing}
\label{subsec:electronics}
\vspace{-1mm}
To support feedback control under sensing constraints representative of ingestible capsule applications, the capsule integrates a compact embedded electronics stack that enables onboard inertial sensing and wireless data transmission. The design targets reduced reliance on high–rate external imaging by combining inertial dead reckoning with intermittent vision-based (or X-ray) pose updates through sensor fusion.

\begin{figure}[t]
  \centering
  \includegraphics[width=\linewidth]{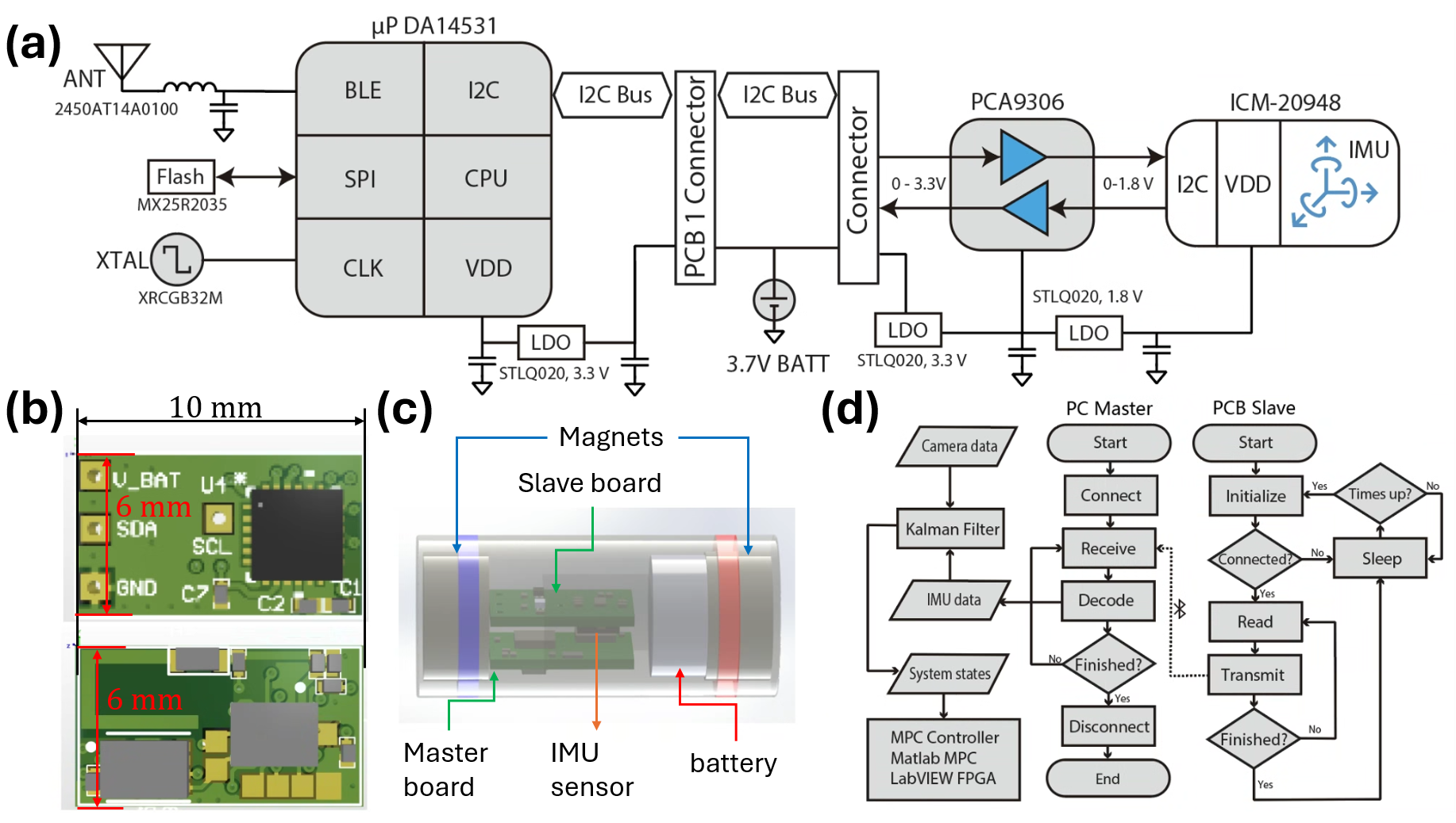}
  \vspace{-8mm}
  \caption{Embedded electronics and inertial sensing architecture. (a) Block diagram of the internal electronics, including the BLE-enabled microcontroller, IMU, level shifter, regulators, and antenna. (b) PCB layouts and dimensions of the master and slave boards. (c) Assembled PCBs and 3D model illustrating integration within the capsule body. (d) Data flow and software architecture for sensor fusion and feedback control, showing IMU and camera data fused via an EKF and supplied to the MPC controller.}
  \label{fig:embedded_electronics}
  \vspace{-5mm}
\end{figure}

\paragraph*{Embedded hardware architecture.}
The internal electronics are realized using two small, modular printed circuit boards (PCBs) designed to fit within the capsule body without disturbing the internal magnet layout or rolling dynamics (Fig.~\ref{fig:embedded_electronics}, (a)-(c)). The \emph{master PCB} hosts a Bluetooth Low Energy (BLE) system-on-chip microcontroller (DA14531, Dialog Semiconductor), which integrates the CPU, BLE radio, and peripheral interfaces with verified Bluetooth signal strength in swine models. The board includes a ceramic antenna (2450AT14A0100, Johanson Technology), a crystal oscillator (XRCGB32M, Murata Manufacturing), a flash memory (MX25R2035, Macronix), and a low-dropout (LDO) voltage regulator (STLQ020, 3.3~V, STMicroelectronics). The microcontroller manages BLE communication while operating in low-power modes when disconnected to extend battery life.

The \emph{slave PCB} carries a nine-axis inertial measurement unit (IMU, ICM-20948, TDK InvenSense), which provides triaxial accelerometer and gyroscope measurements. Since the IMU operates at 1.8~V logic while the microcontroller uses 3.3~V, an I\textsuperscript{2}C bidirectional level shifter (PCA9306, Texas Instruments) is used to interface the two boards. A dedicated 1.8~V LDO regulator (STLQ020, STMicroelectronics) supplies the IMU. The two PCBs are  powered together by a 3.7~V lithium battery.

\paragraph*{Wireless communication and data acquisition.}
IMU data are streamed wirelessly from the capsule to a host computer using BLE. On the host side, a Python-based BLE interface (Bleak) is used to establish the connection, decode IMU packets according to the device datasheet, and forward the measurements to the state estimation pipeline. The DA14531 automatically enters a low-power sleep state when not actively connected, resulting in sub-milliampere average current consumption to extend battery life.

\paragraph*{Sensor fusion and role in feedback control.}
As illustrated in Fig.~\ref{fig:embedded_electronics}(d), inertial measurements from the IMU are fused with vision-based pitch estimates using an extended Kalman filter (EKF). A front-facing camera provides intermittent or reduced-rate pitch measurements that serve as drift-correcting updates, analogous to sparse X-ray snapshots in practical ingestible capsule scenarios. Between these updates, the EKF propagates the capsule orientation using IMU data, providing a continuous estimate of the pitch angle for feedback control.

The fused pitch angle estimates are transmitted to the control computer and used as the measured state in the model predictive controller described in Section~\ref{sec:model_control}. This architecture enables systematic evaluation of the trade-offs between sensing rate, estimation accuracy, and closed-loop performance, and demonstrates the feasibility of maintaining pitch control under constrained external imaging conditions.

\vspace{-4mm}
\subsection{FEA Field Characterization and Torque/Force Mapping}
\label{subsec:fea_map}
\vspace{-1mm}
Finite-element analysis (FEA) in ANSYS Maxwell is used to optimize the actuation system and create the torque/force lookup table later used by the controller. A three-dimensional magnetostatic model explicitly includes the four coil-core assemblies and a single internal magnet placed at the corresponding in-capsule location. The second magnet is omitted during each solve because the mutual magnet--magnet attraction is internal to the rigid capsule assembly and does not contribute to the net external moment about the rolling contact; the field-generated force/torque from the coils on each magnet location is instead computed by evaluating each location consistently within the same external field environment.

The model is embedded in a large air region with a zero-tangential-\emph{H} boundary condition applied on the outer surface to approximate free space. A length-based tetrahedral mesh is used, with maximum element size $1~\mathrm{mm}$ inside a $20~\mathrm{mm}\times 20~\mathrm{mm}\times 20~\mathrm{mm}$ refinement box enclosing the magnet, and a coarser mesh elsewhere to reduce computation.

\begin{figure}[t]
  \centering
  \includegraphics[width=1\linewidth]{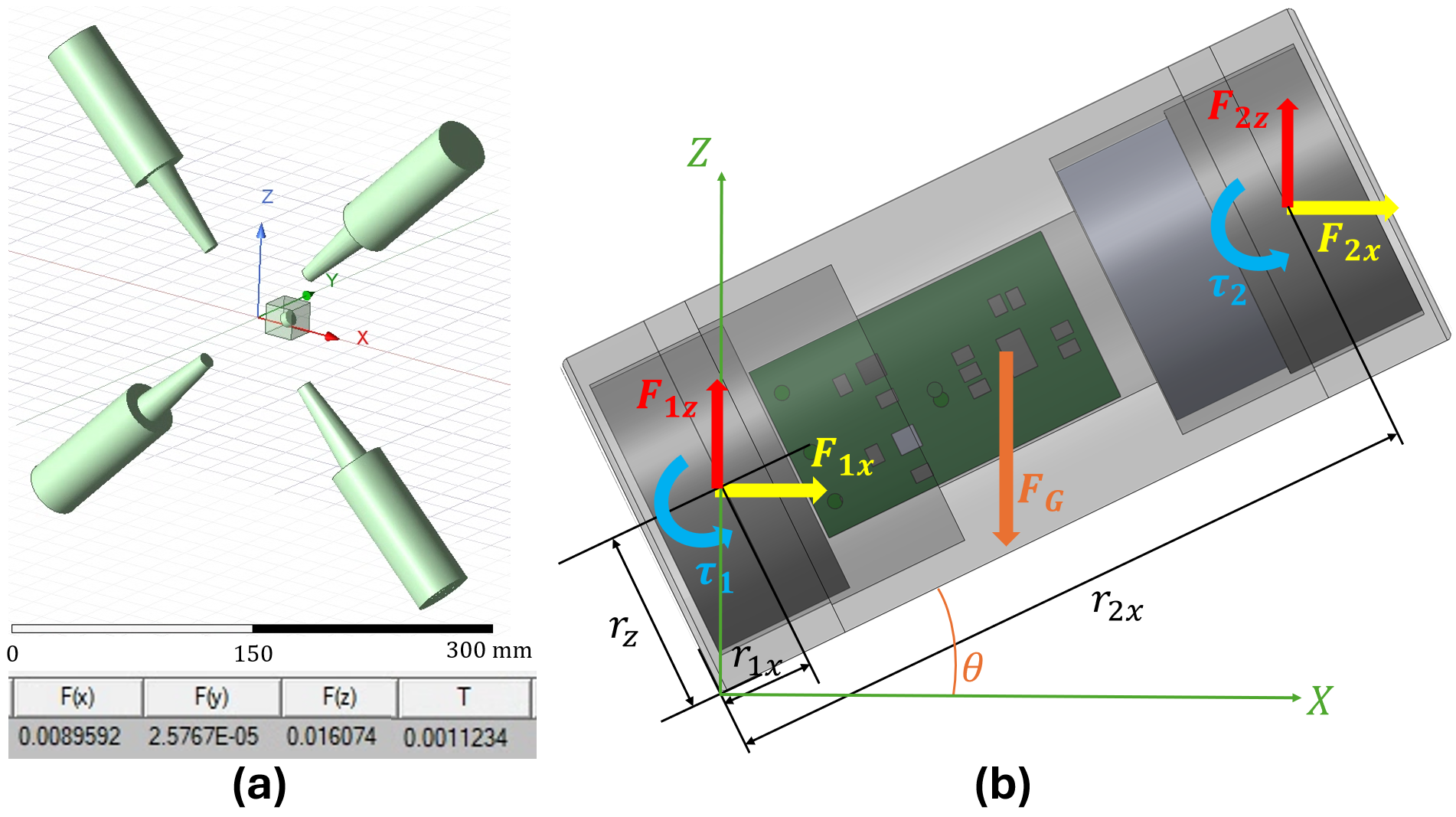}
  \vspace{-8mm}
  \caption{Finite-element magnetic actuation characterization and corresponding free-body diagram for pitch dynamics modeling. (a) Example ANSYS Maxwell magnetostatic solution illustrating the coil–core assemblies, a single internal magnet, and the reported magnetic force components $(F_x,F_y,F_z)$ (newtons) and magnetic torque $T$ (newton-meters) for the vertical actuation mode at $1~\mathrm{A}$ and $\theta=30^\circ$. The near-zero $F_y$ component is consistent with the planar modeling assumption. (b) Free-body diagram of the capsule during pitch motion. Each internal magnet experiences a force $(F_{kx},F_{kz})$ and a direct magnetic torque $\tau_k$ due to the external electromagnetic field. The force and torque quantities obtained from the finite-element simulations in (a) are used to populate the magnetic actuation terms in the pitching dynamics, with the total magnetic moment about the rolling contact point given by the sum of direct torques and $r \times F$ contributions from both magnets.}
  \label{fig:ansys_FBD}
  \vspace{-5mm}
\end{figure}

Two actuation patterns are simulated to match the experimental modes: (i) \emph{diagonal} actuation, where the two coils on one diagonal are driven with $1~\mathrm{A}$ DC current and the other two are off, producing a field whose dominant direction aligns with the diagonal; and (ii) \emph{vertical} actuation, where all four coils are driven with identical $1~\mathrm{A}$ DC currents, producing a predominantly vertical field near the workspace center. For each mode, solutions are computed over a set of discrete capsule pitch angles $\theta_j\in[0^\circ,90^\circ]$ with $5^\circ$ increments. Maxwell reports the magnetic force components $(F_x,F_y,F_z)$ (newtons) and magnetic torque $T$ (newton-meters) on the magnet for each $\theta_j$. An example solution at $\theta=30^\circ$ is illustrated in Fig.~\ref{fig:ansys_FBD}(a). In the symmetric vertical-field mode, $F_y$ is observed to be negligible compared to $F_x$ and $F_z$; this supports the planar modeling assumption used later in the control-oriented dynamics.

The resulting per-ampere force/torque data are exported as lookup tables over $\theta_j$ and are later interpolated to supply the MPC with a continuous actuation model. Details of how these maps enter the prediction model are provided in Section~\ref{sec:model_control}.
\vspace{-3mm}
\subsection{System Integration and Real-Time Implementation}
\label{subsec:integration_realtime}
\begin{figure}[t]
  \centering
  \includegraphics[width=0.47\textwidth]{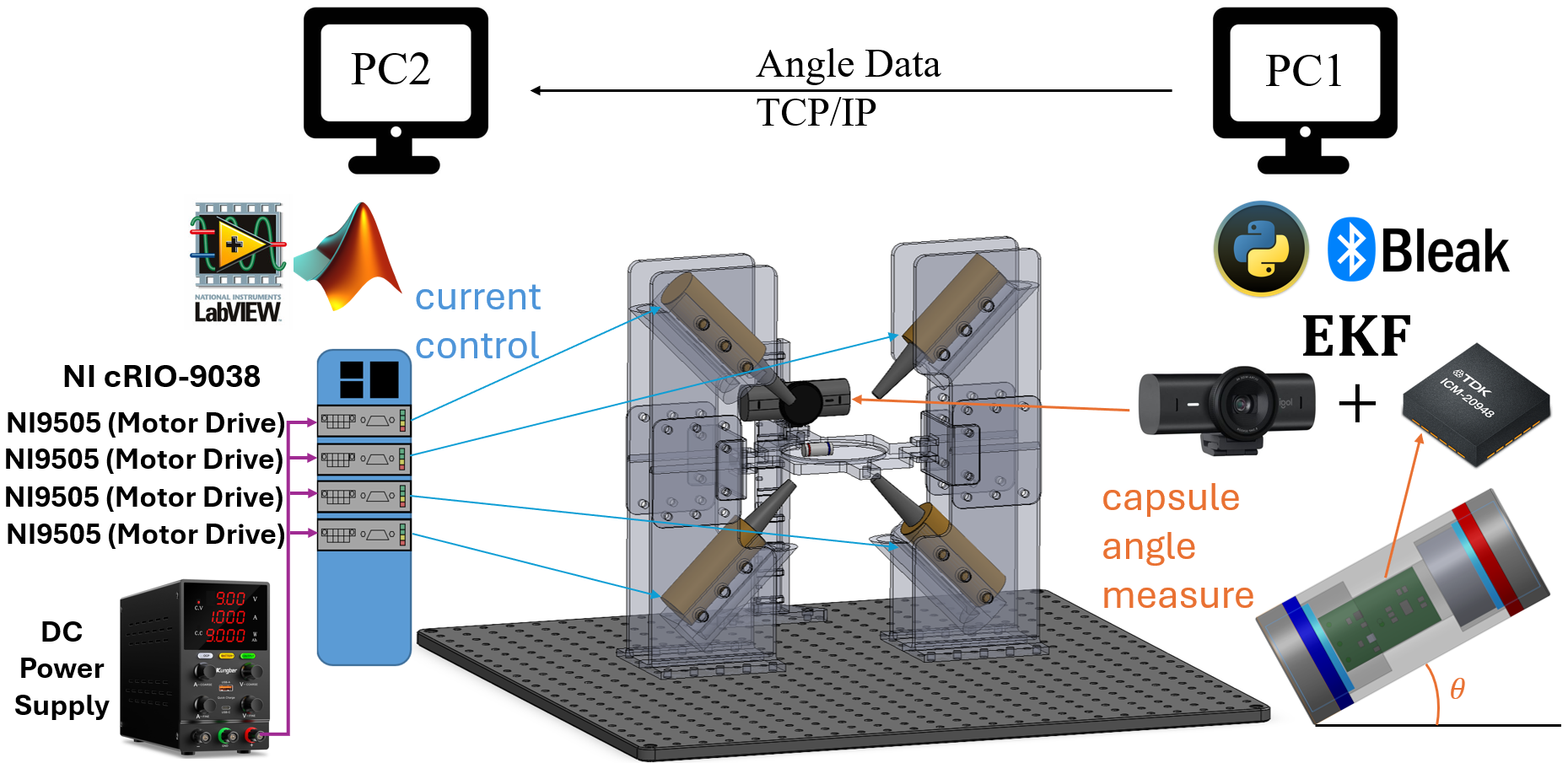}
  \vspace{-3mm}
  \caption{Sensing and real-time control architecture. PC~1 performs
  vision-based pitch estimation (and IMU acquisition/fusion when enabled) and
  sends the pitch estimate to PC~2 via TCP/IP. PC~2 runs LabVIEW, calls the
  MPC via a MATLAB Script Node, and drives each coil through a separate
  NI~9505 current-control module.}
  \label{fig:realtime_arch}
  \vspace{-5mm}
\end{figure}

\vspace{-1mm}
Figure~\ref{fig:realtime_arch} summarizes the sensing and real-time control architecture used in experiments. Two networked computers separate sensing from low-level current control. PC~1 handles vision processing (and BLE IMU data acquisition when enabled) in a Python script, while PC~2 runs LabVIEW for deterministic current actuation and invokes the MPC written in a MATLAB Script Node.

\textbf{Angle sensing (vision and IMU).}
A front-facing camera mounted perpendicular to the silicone surface tracks the colored fiducials and computes a scalar pitch measurement. When inertial sensing is enabled, the on-board IMU streams measurements via BLE to PC~1, which runs the EKF to fuse IMU dead-reckoning with vision updates (vision serving as the intermittent ground-truth correction analogous to reduced-rate X-ray snapshots). The resulting pitch angle estimate is transmitted to PC~2 over TCP/IP.

\textbf{Current control and MPC execution.}
Each coil is driven by a dedicated NI 9505 current-driver module powered by a DC supply and controlled through a CompactRIO-9038. PC~2 receives the latest pitch estimate, updates the state used by the controller, calls the MPC routine via the MATLAB Script Node, and outputs current references to the NI~9505 modules. The driver dynamics and actuator constraints used in the MPC are identified from this setup and are reported in Section \ref{sec:model_control}.

\vspace{-3mm}
\section{Magneto--Mechanical Modeling and Control}
\label{sec:model_control}
\vspace{-1mm}
This section presents a unified control-oriented formulation for magnetic pitch regulation of the capsule robot. The section first develops a lumped magneto-mechanical model that maps coil current to capsule pitch dynamics using finite-element  magnetic characterization. This model is then embedded into a constrained MPC framework. Finally, a state estimation and sensing subsection describes the multi-rate measurement architecture and its integration via an extended Kalman filter.

\vspace{-4mm}
\subsection{Magneto-Mechanical Modeling for Pitch Dynamics}
\label{subsec:model_pitch}
\vspace{-1mm}
We consider planar pitch motion of the capsule in the sagittal plane, as illustrated in Fig.~\ref{fig:ansys_FBD}(b). A world frame $\{X,Z\}$ is fixed to the compliant silicone surface, with $X$ tangential and $Z$ normal to the surface. The capsule is modeled as a rigid cylinder of mass $m=7.42~\mathrm{g}$ rolling without slip about a contact point. The pitch angle $\theta \in [0,\pi/2]$ is measured counterclockwise from the horizontal.

Two cylindrical permanent magnets are embedded symmetrically inside the capsule and rigidly attached to the body. In the body-fixed frame $\{\bar{x},\bar{z}\}$, the position of magnet $k\in\{1,2\}$ relative to the rolling contact point is

\vspace{-4.5mm}

\begin{equation}
    \bar{\mathbf{r}}_k = [\, r_{kx} \;\; 0 \;\; r_{kz} \,]^T,
\end{equation}
with the corresponding world-frame position

\vspace{-4mm}

\begin{equation}
    \mathbf{r}_k(\theta) = \mathbf{R}(\theta)\,\bar{\mathbf{r}}_k,
\end{equation}

\vspace{-2mm}
\noindent where $\mathbf{R}(\theta)$ denotes the planar rotation matrix.

Each magnet experiences a magnetic force

\vspace{-4mm}

\begin{equation}
    \mathbf{F}_k(\theta,i)
    =
    [\, F_{kx}(\theta,i) \;\; 0 \;\; F_{kz}(\theta,i) \,]^T
\end{equation}

\vspace{-2mm}

\noindent and a direct magnetic torque $\tau_k(\theta,i)$ due to the externally generated magnetic field. Mutual forces between the two internal magnets are internal to the rigid capsule and therefore do not contribute to the external torque balance.

Rather than relying on closed-form magnetic models, the force components $F_{kx}(\theta,i)$, $F_{kz}(\theta,i)$ and the direct torque $\tau_k(\theta,i)$ are obtained from finite-element magnetostatic simulations performed in ANSYS Maxwell, as described in Section~\ref{subsec:fea_map}. These simulations are conducted at a unit current, yielding per-ampere force and torque data over a discrete set of capsule pitch angles. The resulting data are stored as lookup tables indexed by $\theta$ and interpolated online to provide a continuous actuation model for control.

The magnetic forces acting at offset locations within the capsule generate additional moments about the rolling contact point. For magnet $k$, this contribution is given by

\vspace{-6mm}

\begin{equation}
    \tau_{F_k}(\theta)
    =
    \big( \mathbf{r}_k(\theta) \times \mathbf{F}_k(\theta) \big)_y
    =
    r_{kx}(\theta) F_{kz}(\theta)
    - r_{kz}(\theta) F_{kx}(\theta).
\end{equation}

\vspace{-1.5mm}

The total magnetic torque available at the capsule is expressed in control-oriented form as

\vspace{-4mm}

\begin{equation}
    \tau_{\mathrm{mag}}(\theta,i)
    =
    i\,\tau_{\mathrm{FE}}(\theta),
    \label{eq:magnetic_torque_map}
\end{equation}

\vspace{-2.5mm}

\noindent where the torque-per-ampere map

\vspace{-3.5mm}

\begin{equation}
    \tau_{\mathrm{FE}}(\theta)
    =
    \alpha
    \Big(
    \tau_1(\theta) + \tau_2(\theta)
    + \tau_{F_1}(\theta) + \tau_{F_2}(\theta)
    \Big)
\end{equation}

\vspace{-2mm}

\noindent is obtained by interpolating the finite-element lookup tables. The scalar $\alpha$ accounts for uncertainties in magnet remanence, manufacturing tolerances, and modeling idealizations.

The pitch dynamics is then described by

\vspace{-5mm}

\begin{equation}
    I_p \ddot{\theta}
    =
    \tau_{\mathrm{mag}}(\theta,i)
    - m g L \cos\theta,
\end{equation}

\vspace{-2mm}
\noindent where $I_p$ denotes the moment of inertia about the contact point and $L$ is the distance from the contact to the capsule center of mass. This nonlinear, angle-dependent dynamic model forms the basic model used in the predictive controller.

\vspace{-4mm}
\subsection{Model Predictive Control Formulation}
\vspace{-1mm}
The magneto--mechanical model is embedded into a constrained MPC framework to regulate the capsule pitch angle. The control-oriented state vector is defined as

\vspace{-4mm}

\begin{equation}
    \mathbf{x}_k = [\, \theta_k \;\; \omega_k \;\; i_k \,]^T,
\end{equation}

\vspace{-2mm}

\noindent where $\omega_k = \dot{\theta}_k$ and $i_k$ denotes the actual coil current.

The discrete-time prediction model takes the form

\vspace{-4mm}

\begin{equation}
    \mathbf{x}_{k+1}
    =
    \mathbf{A}_k \mathbf{x}_k
    +
    \mathbf{B}_k u_{\mathrm{cmd},k}
    +
    \mathbf{d}_k,
\end{equation}

\vspace{-2mm}

\noindent where $u_{\mathrm{cmd}}$ is the current command issued by the MPC. The matrices $\mathbf{A}_k$ and $\mathbf{B}_k$ are updated online to reflect the angle-dependent magnetic torque map.

The NI~9505 current driver dynamics are approximated as a first-order system,

\vspace{-4mm}

\begin{equation}
    \dot{i} = \frac{1}{\tau_c}(u_{\mathrm{cmd}} - i),
\end{equation}

\vspace{-2mm}

\noindent and discretized using the MPC sampling period $T_s$.

At each control step, the MPC solves a finite-horizon quadratic program to compute the optimal command sequence

\vspace{-4mm}

\begin{equation}
    \mathbf{U}
    =
    [\, u_{\mathrm{cmd},k} \;\; \dots \;\; u_{\mathrm{cmd},k+N_p-1} \,]^T.
\end{equation}

\vspace{-2mm}

The cost function penalizes pitch tracking error, angular velocity, input magnitude, and input variation:

\vspace{-5mm}

\begin{align}
    J
    =
    \sum_{j=1}^{N_p}
    \Big(
    Q_\theta (\theta_{k+j}-\theta_{\mathrm{ref}})^2
    + Q_\omega \omega_{k+j}^2
    \Big)\nonumber\\
    +
    \sum_{j=0}^{N_p-1} R u_{\mathrm{cmd},k+j}^2
    +
    \sum_{j=0}^{N_p-2} S \Delta u_{k+j}^2.
\end{align}

\vspace{-2mm}

Amplitude constraints $|u_{\mathrm{cmd}}| \le 1~\mathrm{A}$ and slew-rate constraints $|\Delta u| \le \Delta u_{\max}$ are enforced explicitly, yielding a convex quadratic program that is solved in real time.

\vspace{-4mm}

\subsection{State Estimation and Multi-Rate Sensing Architecture}
\begin{figure}[t]
  \centering
  \includegraphics[width=\linewidth]{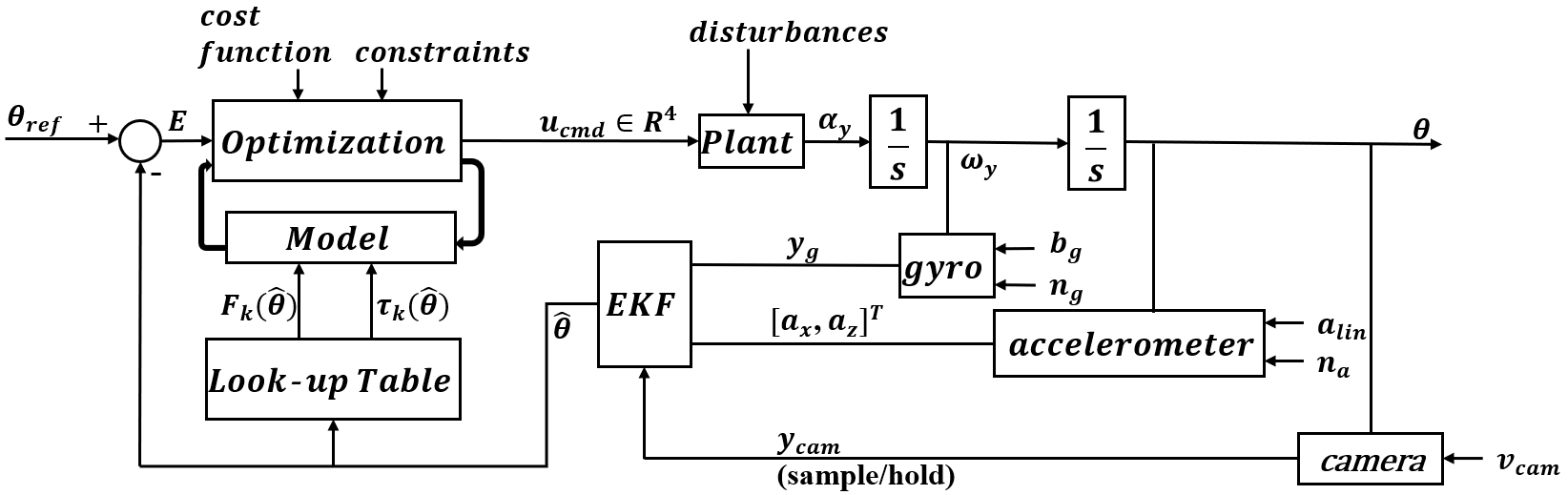}
  \vspace{-8mm}
  \caption{Closed-loop structure of the pitch–angle control system with multi-rate sensing and state estimation. The MPC receives the reference angle and the estimated pitch angle $\hat{\theta}$ from an extended Kalman filter (EKF), solves a constrained quadratic program, and outputs a vector-valued current command $\mathbf{u}_{\mathrm{cmd}}\in\mathbb{R}^4$ to the actuation plant (current drivers, electromagnetic coils, and capsule). The EKF propagates the pitch state using high-rate inertial measurements from the onboard IMU (50~Hz gyroscope and accelerometer), with accelerometer signals providing gravity-referenced orientation cues subject to motion-induced disturbances during rolling. Low-rate camera-based pitch observations, sampled at $1~\mathrm{Hz}$, provide intermittent absolute corrections that suppress drift and yield a continuous pitch estimate for feedback control.}
  \label{fig:mpc_block}
  \vspace{-5mm}
\end{figure}

\vspace{-1mm}
The closed-loop sensing–estimation–control architecture is illustrated in Fig.~\ref{fig:mpc_block}. To support reliable pitch regulation under intermittent vision, state estimation leverages high-rate onboard inertial sensing for continuous propagation and sparse vision measurements for drift correction.

The capsule integrates an inertial measurement unit (IMU) that provides triaxial gyroscope and accelerometer measurements at $50~\mathrm{Hz}$. The gyroscope supplies the measured pitch-axis angular velocity

\vspace{-5mm}

\begin{equation}
y_g = \omega_y + b_g + n_g,
\end{equation}

\vspace{-2mm}

\noindent where $b_g$ is a slowly varying bias and $n_g$ denotes measurement noise. While gyroscope integration enables high-rate pitch angle propagation, gyro-only dead-reckoning exhibits significant drift over time due to bias accumulation, motivating the incorporation of additional orientation cues.

The accelerometer outputs $(a_x,a_z)$ represent specific acceleration components in the capsule body frame. Under the rolling-without-slip assumption adopted in this work, the measured signal is dominated by gravity projection, with additional non-gravitational contributions arising from motion-induced effects associated with rigid-body kinematics and surface contact. Because these translational accelerations remain small relative to gravity in the operating regime considered, the accelerometer provides a nonlinear, gravity-referenced observation of the capsule pitch angle. Accordingly, the accelerometer measurement model is expressed as

\vspace{-2mm}

\begin{equation}
\begin{bmatrix} a_x \\ a_z \end{bmatrix}
=
g
\begin{bmatrix}
\sin\theta \\
\cos\theta
\end{bmatrix}
+
\mathbf{a}_{\mathrm{lin}} + \mathbf{n}_a,
\end{equation}

\vspace{-1mm}

\noindent where $\mathbf{a}_{\mathrm{lin}}$ represents translational specific force induced by rolling dynamics (e.g., tangential and centripetal components), and $\mathbf{n}_a$ captures sensor noise and unmodeled effects. In the operating regime considered here, $\|\mathbf{a}_{\mathrm{lin}}\|$ remains small relative to $g$, allowing the accelerometer to provide a reliable, low-frequency orientation constraint despite contamination from motion-induced acceleration.

A vision-based module provides direct pitch angle observations extracted from camera images. Although images are acquired at approximately $30~\mathrm{fps}$, the resulting observation

\vspace{-4mm}

\begin{equation}
y_{\mathrm{cam}} = \theta + v_{\mathrm{cam}}
\end{equation}

\vspace{-2mm}

\noindent is intentionally downsampled and held at $1~\mathrm{Hz}$ using a sample-and-hold scheme as an accurate low-rate observation for sensor fusion as described later. This design emulates intermittent imaging conditions relevant to ingestible capsule applications and enables systematic evaluation of estimator robustness under constrained external sensing.

An extended Kalman filter (EKF) fuses the gyroscope, accelerometer, and low-rate camera measurements to estimate the capsule pitch angle. The EKF state is defined as

\vspace{-4mm}

\begin{equation}
\mathbf{x} = \begin{bmatrix} \theta & b_g \end{bmatrix}^\top,
\end{equation}

\vspace{-2mm}

\noindent and is propagated at the IMU rate according to

\vspace{-6mm}

\begin{equation}
\theta_{k+1} = \theta_k + (y_{g,k} - b_{g,k})\Delta t + w_{\theta,k}, \quad
b_{g,k+1} = b_{g,k} + w_{b,k},
\end{equation}

\vspace{-1mm}

\noindent with $\Delta t = 1/50~\mathrm{s}$. Accelerometer measurements provide correction through the gravity projection model, while camera measurements ($1~\mathrm{Hz}$) serve as intermittent absolute updates that realign the estimate and suppress long-term drift.

The resulting pitch estimate $\hat{\theta}$ is supplied to the MPC as the feedback state. The MPC update rate is constrained by real-time quadratic program (QP) solution time, approximately $100$--$150~\mathrm{ms}$ per iteration, yielding a control update rate of $6$--$10~\mathrm{Hz}$. The EKF thus serves as a multi-rate interface between high-bandwidth inertial sensing, intermittent vision, and optimization-based control.

Overall, this sensing and estimation strategy enables accurate pitch regulation with reduced dependence on high-rate imaging, while remaining compatible with the computational and sensing constraints of ingestible capsule robotic systems.




\vspace{-4mm}
\section{Experimental Validation}
\label{sec:experiments}
\vspace{-1mm}
\subsection{Experimental Setup and Protocol}
\begin{figure}[t]
  \centering
  \includegraphics[width=0.6\linewidth]{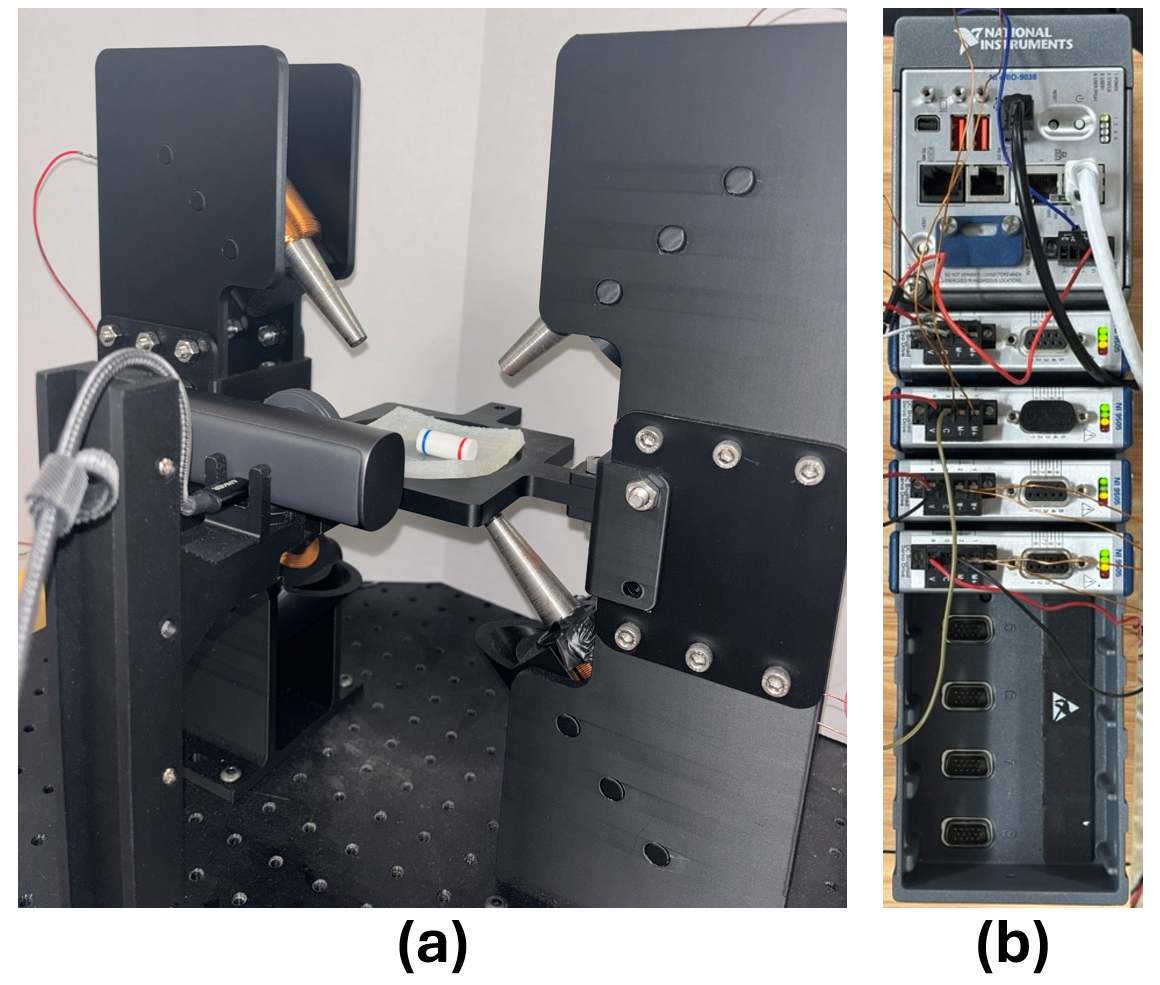}
  \vspace{-4mm}
  \caption{Experimental setup for magnetic pitch control. (a) Benchtop platform with four external electromagnetic coils actuating a capsule robot rolling on a compliant, stomach-inspired silicone surface. A front-facing camera provides vision-based pitch measurements. (b) CompactRIO-9038 chassis populated with four NI~9505 current-control modules. Each module is powered by a dedicated $9~\mathrm{V}$ DC supply and independently drives a single coil.}
  \label{fig:exp_setup}
  \vspace{-5mm}
\end{figure}

All experiments are conducted on the benchtop platform shown in Fig.~\ref{fig:exp_setup}. As shown in Fig.~\ref{fig:exp_setup}(a), four external electromagnetic coils surround a compliant, stomach-inspired silicone surface on which the capsule robot rolls. The silicone test surface is 3D printed and has a Shore hardness of 18, selected to mimic the compliance of the gastric wall while providing repeatable contact conditions. A front-facing camera is mounted normal to the surface to provide vision-based pitch measurements during experiments.

The electromagnetic coils are driven by four independent NI~9505 current-control modules mounted in a CompactRIO-9038 chassis, shown in Fig.~\ref{fig:exp_setup}(b). Each NI~9505 module is powered by a dedicated DC power supply set to a constant voltage of $9~\mathrm{V}$ and is connected to a single coil, enabling independent current control across all four actuators. Coil currents are commanded by the MPC through LabVIEW running on PC~2, which interfaces with the CompactRIO in a deterministic real-time loop.

Vision processing and inertial data acquisition are handled on PC~1. Camera images are processed to extract pitch measurements, while onboard IMU data are streamed wirelessly via BLE. When enabled, an extended Kalman filter (EKF) runs on PC~1 to fuse inertial measurements with intermittent camera updates and transmits the estimated pitch state to PC~2 via TCP/IP for feedback control.

The closed-loop update rate is limited primarily by real-time computation rather than sensor bandwidth. In particular, the MPC loop (LabVIEW $\rightarrow$ MATLAB Script Node QP solve $\rightarrow$ current command) runs at approximately $100$–$150~\mathrm{ms}$ per iteration, corresponding to an effective control rate of $6$–$10~\mathrm{Hz}$. Each trial is terminated once the capsule reaches a quasi–steady state, defined as sustained oscillation within a $\pm 2.5^\circ$ band about the target pitch angle, consistent with the settling-time criterion introduced earlier. As a result, trial durations vary across control strategies and initial conditions. All pitch trajectories are recorded for post-processing and quantitative comparison.

\vspace{-5mm}
\subsection{Test Cases and Compared Control/Sensing Configurations}
\label{subsec:exp_cases}
\vspace{-1mm}
The target pitch angle of $30^\circ$ is selected to represent a practically challenging docking configuration on the compliant surface. At this inclination, the capsule cannot reliably maintain contact through gravity alone, yet stable contact is required for contact-rich tasks such as localized drug delivery or system identification. 

To enable quantitative comparison, settling time is defined as the time required for the pitch angle to enter and remain within a $\pm2.5^\circ$ band around the $30^\circ$ target. This criterion captures both convergence speed and residual oscillatory behavior, and is particularly appropriate for the on--off baseline, which exhibits an approximately exponential decay envelope.

Two maneuvers are evaluated: (1) Horizontal-to-inclined: $\;0^\circ \rightarrow 30^\circ$ and (2) Upright-to-inclined: $\;90^\circ \rightarrow 30^\circ$.

The second maneuver is more challenging because the capsule begins near an upright configuration where the magnetic torque–angle relationship and contact mechanics can produce large transients. To move the capsule away from this near-vertical equilibrium, a brief open-loop actuation is first applied by energizing the diagonal coil pair at $0.3~\mathrm{A}$ for $0.15~\mathrm{s}$. After this initial tilt, the feedback controller is engaged. This pre-actuation appears as the initial transient in the measurement.

We compare the following strategies across both maneuvers:

\paragraph*{(A) On--off control (baseline).}
The diagonal coils are switched on to a fixed current until the capsule reaches a quasi-steady orientation. This method represents a simple bang-bang actuation without input shaping or feedback stabilization.

\paragraph*{(B) MPC with camera-only feedback at 30~Hz.}
The MPC receives pitch measurements obtained from the camera at full frame rate (30~Hz acquisition). At each MPC step, the most recent measurement is used as the feedback state.

\paragraph*{(C) MPC with camera-only feedback limited to 5~Hz.}
To emulate reduced-rate imaging, the camera acquisition is throttled to 5~Hz and only those measurements are provided as feedback. This case tests whether low-rate vision alone is sufficient for stable closed-loop pitch regulation.

\paragraph*{(D) MPC with camera--IMU fusion (camera 1~Hz sample/hold).}
To emulate intermittent clinical imaging, camera-based pitch observations are downsampled and held at 1~Hz, while the IMU provides 50~Hz inertial measurements. An EKF fuses these modalities to produce a continuous pitch estimate $\hat{\theta}$, which is provided to the MPC at each control update.

\vspace{-4mm}

\subsection{Results: $0^\circ \rightarrow 30^\circ$ Reorientation}
\label{subsec:exp_0to30}
\begin{figure}[t]
  \centering
  \includegraphics[width=1\linewidth]{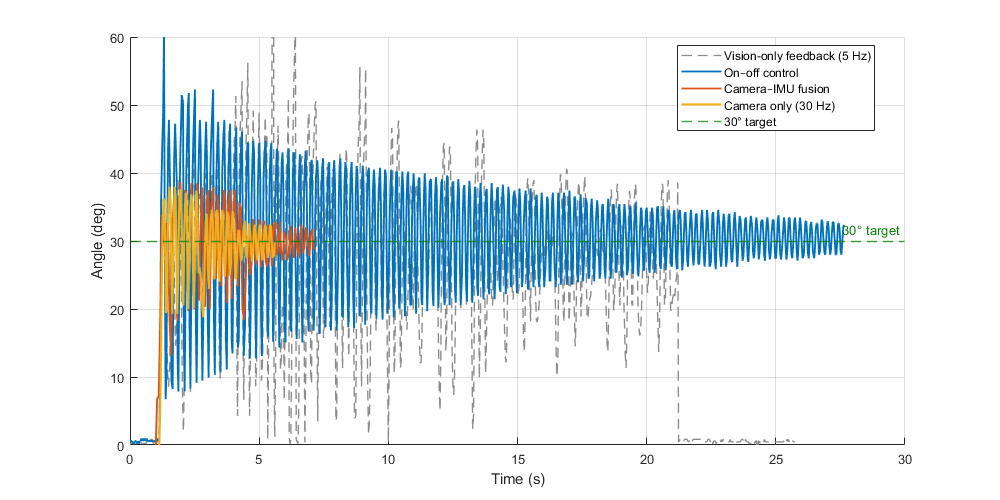}
  \vspace{-9mm}
  \caption{Measured pitch trajectories for the $0^\circ \rightarrow 30^\circ$ maneuver comparing: on--off baseline, MPC with camera-only feedback at 30~Hz, MPC with camera-only feedback limited to 5~Hz, and MPC with camera--IMU fusion using 1~Hz camera sample/hold. Control actuation is initiated at $t = 1~\mathrm{s}$ for all cases. Settling time is evaluated using a $\pm2.5^\circ$ band around the $30^\circ$ target, highlighting the substantially faster convergence of MPC-based strategies compared to on--off control.}
  \vspace{-8mm}
  \label{fig:angle_0to30}
\end{figure}

\vspace{-1mm}
Figure~\ref{fig:angle_0to30} compares pitch trajectories for the horizontal-to-inclined maneuver under the four control and sensing configurations. Using the $\pm2.5^\circ$ settling criterion defined earlier, clear performance differences emerge across strategies.

MPC with camera-only feedback at 30~Hz achieves the fastest convergence, reaching the target pitch within approximately $4~\mathrm{s}$ with minimal residual oscillation. When camera measurements are fused with onboard IMU data and downsampled to 1~Hz, convergence remains stable but slower, with settling occurring at approximately $6~\mathrm{s}$. This increase in settling time reflects estimator latency and reduced absolute correction rate under intermittent visual updates.

By contrast, the on--off baseline exhibits large overshoot and sustained oscillations, requiring more than $20~\mathrm{s}$ to settle within the target band. Although the oscillation envelope decays gradually, the lack of input shaping and feedback stabilization leads to prolonged convergence. MPC with camera-only feedback limited to 5~Hz becomes visibly unstable, with large excursions and intermittent loss of regulation, and does not consistently achieve settling within the trial duration.

Across all stable cases, residual oscillations persist after convergence. As discussed in Section~\ref{sec:model_control}, this behavior follows directly from the magnetic torque structure in equation~\eqref{eq:magnetic_torque_map}, where corrective torque reverses sign following overshoot. Because the capsule--surface interaction exhibits low inherent damping, these corrections are weakly dissipated, resulting in lightly damped oscillatory motion even under MPC regulation.

\vspace{-3mm}
\subsection{Results: $90^\circ \rightarrow 30^\circ$ Reorientation}
\label{subsec:exp_90to30}
\begin{figure}[t]
  \centering
  \includegraphics[width=1\linewidth]{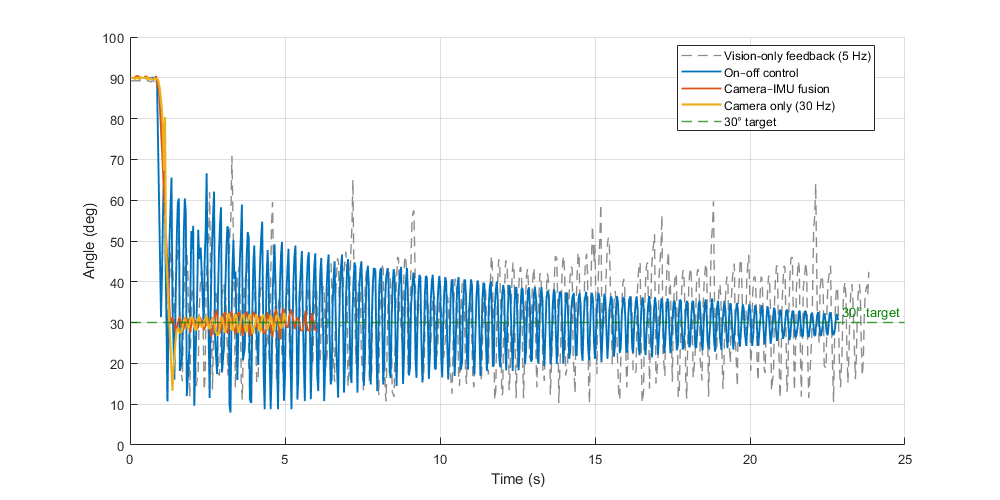}
  \vspace{-9mm}
  \caption{Measured pitch trajectories for the $90^\circ \rightarrow 30^\circ$ maneuver under the same four control and sensing strategies as Fig.~\ref{fig:angle_0to30}. Control actuation is initiated at $t = 1~\mathrm{s}$ following a brief open-loop diagonal-coil actuation used to tilt the capsule away from the upright configuration. Settling time is evaluated using a $\pm2.5^\circ$ band around the $30^\circ$ target, illustrating faster convergence for MPC-based controllers relative to on--off control.}
  \vspace{-5mm}
  \label{fig:angle_90to30}
\end{figure}

\vspace{-1mm}
Figure~\ref{fig:angle_90to30} presents results for the more challenging upright-to-inclined maneuver. In all trials, a brief open-loop diagonal-coil actuation ($0.3~\mathrm{A}$ for $0.15~\mathrm{s}$) is applied to move the capsule away from its near-vertical equilibrium before entering the feedback loop.

After this initial transient, MPC with camera-only feedback at 30~Hz again achieves rapid and stable convergence, reaching the target pitch in approximately less than $4~\mathrm{s}$. MPC with camera--IMU fusion settles slightly more slowly, achieving convergence within approximately $5~\mathrm{s}$ while maintaining stable regulation despite the 1~Hz camera update rate. These results demonstrate that estimator-assisted feedback preserves fast reorientation even from steep initial configurations.

In contrast, the on--off baseline exhibits the slowest convergence and largest oscillations, requiring more than $20~\mathrm{s}$ to settle. As in the $0^\circ \rightarrow 30^\circ$ case, camera-only feedback at 5~Hz produces erratic behavior and fails to stabilize the system.

Overall, the relative ordering of convergence speed and stability across control strategies is consistent between shallow and steep initial conditions, underscoring the robustness of the finite-element-informed MPC framework and the effectiveness of inertial fusion under intermittent imaging.

\vspace{-4mm}
\section{Discussion}
\label{sec:discussion}
\vspace{-1mm}
The experiments support two main conclusions. First, model-based feedback control substantially improves pitch reorientation performance compared to simple on-off actuation, yielding faster and more controlled convergence. Second, fusing onboard inertial sensing with intermittent camera measurements enables stable closed-loop pitch control under reduced imaging rates, which is critical for medically constrained sensing scenarios.

\vspace{-4mm}

\subsection{Benefits of MPC-Based Pitch Regulation}
\vspace{-1mm}
Across both maneuvers, on--off control exhibits the largest overshoot and the slowest decay of oscillations. This behavior arises because a fixed current step injects energy without accounting for the nonlinear, angle-dependent magnetic actuation, gravity-induced restoring torque, or contact-induced damping. In contrast, MPC explicitly shapes the current trajectory using a prediction model and a quadratic cost that penalizes pitch error, angular velocity, and input variation while enforcing current amplitude and slew-rate constraints.

\textbf{As a result, MPC achieves substantially faster convergence}: for both the $0^\circ \rightarrow 30^\circ$ and $90^\circ \rightarrow 30^\circ$ maneuvers, MPC-based controllers reach the $\pm2.5^\circ$ settling band within approximately $4$--$6~\mathrm{s}$, whereas the on--off baseline typically requires more than $20~\mathrm{s}$. This represents an improvement in settling time on the order of three to five times, depending on the initial configuration. These results indicate that the primary source of performance improvement is the use of model-based prediction to shape the current trajectory, rather than the application of larger or more abrupt control inputs.

\vspace{-4mm}

\subsection{Impact of Measurement Rate on Closed-Loop Stability}
\vspace{-1mm}
A key observation is that camera-only MPC remains stable with full-rate vision feedback (30~Hz), but becomes unsteady when the camera rate is reduced to 5~Hz. This degradation is not simply due to reduced measurement density, but rather to a mismatch between the MPC update rate (approximately $6$--$10~\mathrm{Hz}$) and the information rate of the feedback signal. When multiple control updates occur between successive measurements, the controller repeatedly acts on stale state information, effectively introducing additional delay and phase lag. In a lightly damped nonlinear pitching system, this can lead to oscillatory amplification and loss of stability, as observed in the low-rate vision-only case.

\vspace{-4mm}

\subsection{Role of Camera--IMU Fusion Under Intermittent Imaging}
\vspace{-1mm}
Camera--IMU fusion provides a practical remedy to this limitation. By downsampling camera measurements to 1~Hz and using the onboard IMU to propagate the attitude estimate between updates, the EKF supplies a continuous pitch estimate to the MPC despite infrequent absolute measurements. Although the fused-estimate response exhibits a longer rise time and slightly increased overshoot compared to full-rate vision feedback, the closed loop remains stable and convergent.

\textbf{Importantly, this stability is achieved with only a modest increase in settling time (on the order of $1$--$2~\mathrm{s}$) relative to full-rate camera feedback, while still outperforming on--off control by a wide margin.} This contrast highlights that onboard inertial sensing is essential for reliable control when external imaging must be sparse.

\vspace{-4mm}

\subsection{Limitations}
\vspace{-1mm}
The current implementation is primarily limited by real-time computation and multi-rate integration. MPC performance is bounded by the QP solution time ($\sim100$--$150~\mathrm{ms}$ per step), while estimator performance under 1~Hz camera updates is influenced by inertial bias modeling and filter latency. Addressing these limitations through faster solvers, improved estimator models, and tighter co-design of estimation and control is expected to further improve transient performance under intermittent sensing.

Overall, the results demonstrate that finite-element-informed MPC enables smooth, rapid, and robust pitch regulation, and that camera--IMU fusion is a key enabler for operating under imaging constraints relevant to practical ingestible capsule applications.

\vspace{-4mm}

\section{Conclusion and Future Work}
\label{sec:conclusion}
\vspace{-1mm}
This work presented a finite-element-informed model predictive control framework for magnetic pitch regulation of an ingestible capsule robot on a compliant, stomach-inspired surface. Angle-dependent magnetic force and torque maps were obtained via three-dimensional finite-element simulations and embedded in a control-oriented rigid-body pitching model that accounts for gravity, rolling contact kinematics, and actuator dynamics. A constrained MPC controller was implemented in real time and experimentally validated under multiple sensing and initialization conditions.

Experiments show that model-based pitch regulation substantially outperforms on--off actuation. For both $0^\circ \rightarrow 30^\circ$ and $90^\circ \rightarrow 30^\circ$ reorientation maneuvers, MPC-based control achieves settling within a $\pm2.5^\circ$ target band in approximately $4$--$6~\mathrm{s}$, compared to more than $20~\mathrm{s}$ for on--off control, corresponding to a three- to five-fold reduction in settling time with reduced oscillatory motion. In addition, experiments show that fusing onboard inertial sensing with intermittent camera measurements via an EKF enables stable closed-loop pitch control even when camera updates are reduced from $30~\mathrm{Hz}$ to $1~\mathrm{Hz}$, supporting operation under medically constrained imaging conditions. These improvements are achieved without increasing actuation magnitude, but instead through predictive shaping of the coil current trajectory based on the nonlinear magneto--mechanical model.

Future work will extend the proposed framework in several directions. On the control side, faster optimization solvers and tailored MPC formulations will be explored to increase control bandwidth and improve transient performance. Robust and adaptive extensions will be investigated to better handle contact uncertainty and variability in surface compliance. On the sensing side, further refinement of inertial–vision fusion, including improved bias modeling and asynchronous update handling, will target reliable operation under even sparser imaging. At the system level, scaling the magnetic actuation workspace and integrating pitch regulation with roll and yaw control will enable coordinated multi-degree-of-freedom locomotion over more complex gastric geometries, supporting future contact-rich diagnostic and therapeutic tasks. A customized end-effector mounted on a robotic arm is being designed for this application.

\section{Acknowledgement}
\label{sec:acknowledgement}
The authors would like to acknowledge UT Austin for its start-up funding to support this research and Ian L. Heyman for initial technical support on Ansys Maxwell simulation.
\vspace{-2mm}

\bibliographystyle{ieeetr}
\bibliography{reference}


\begin{IEEEbiography}[{\includegraphics[width=1in,height=1.25in,clip,keepaspectratio]{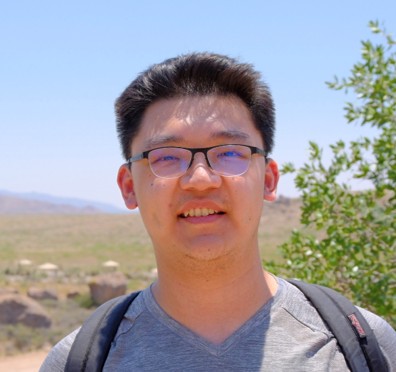}}]{Chongxun Wang} received his B.S. degree in mechanical engineering from The Pennsylvania State University in 2021, and M.S. degree in mechanical engineering from Stanford University in 2023. He is currently pursuing his Ph.D. degree in mechanical engineering at The University of Texas at Austin where he joined in 2024. His research interests include mechanical system design and control of dynamical systems, with a focus on locomotion of capsule robots.
\end{IEEEbiography}

\begin{IEEEbiography}[{\includegraphics[width=1in,height=1.25in,clip,keepaspectratio]{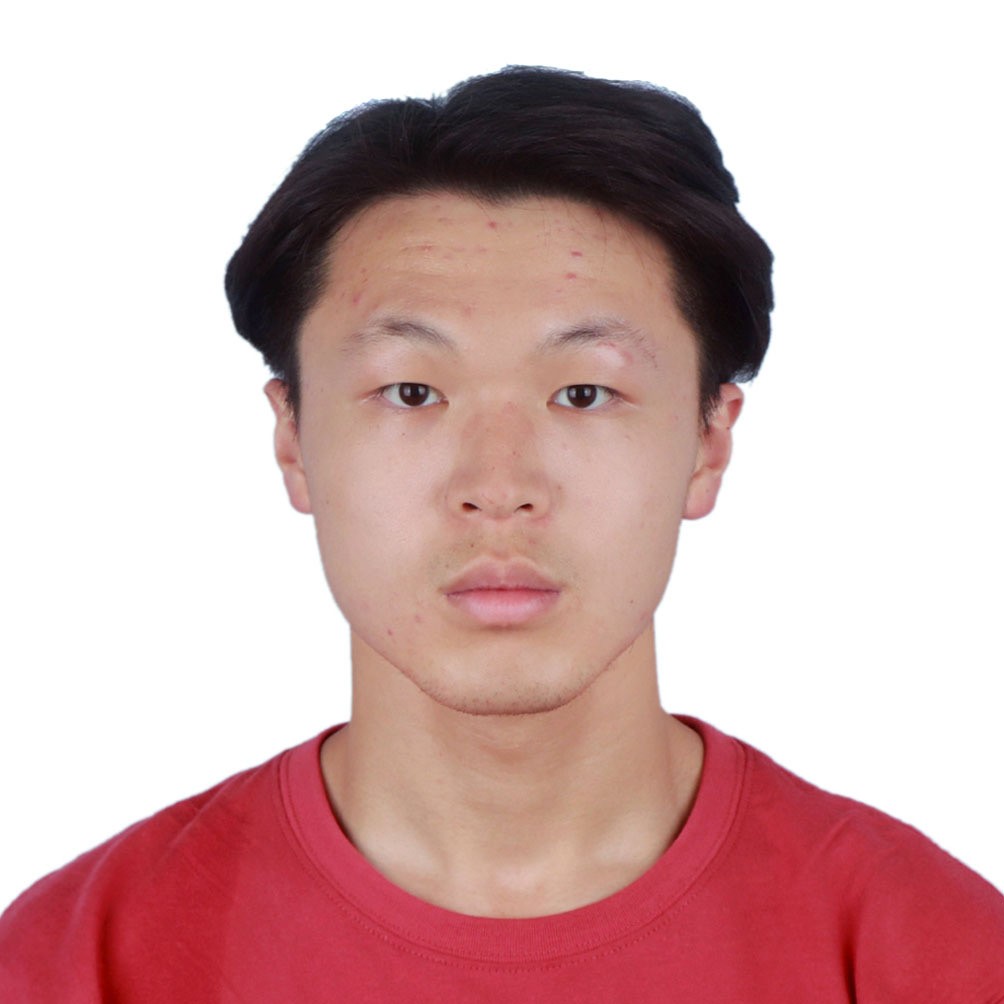}}]{Zikang Shen} is currently pursuing his B.S degree in mechanical engineering at the University of Texas at Austin. His academic interests focus on dynamic systems and controls.
\end{IEEEbiography}

\begin{IEEEbiography}[{\includegraphics[width=1in,height=1.25in,clip,keepaspectratio]{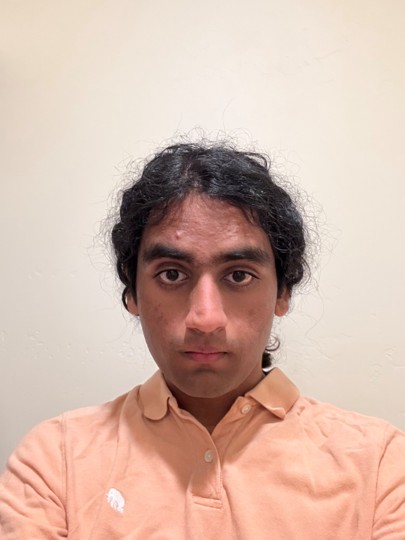}}]{Apoorav Rathore} is pursuing his B. S. in Electrical \& Computer Engineering at the University of Texas at Austin. His academic interests include Embedded Systems, Software Development, and Computer Architecture. 
\end{IEEEbiography}

\begin{IEEEbiography}[{\includegraphics[width=1in,height=1.25in,clip,keepaspectratio]{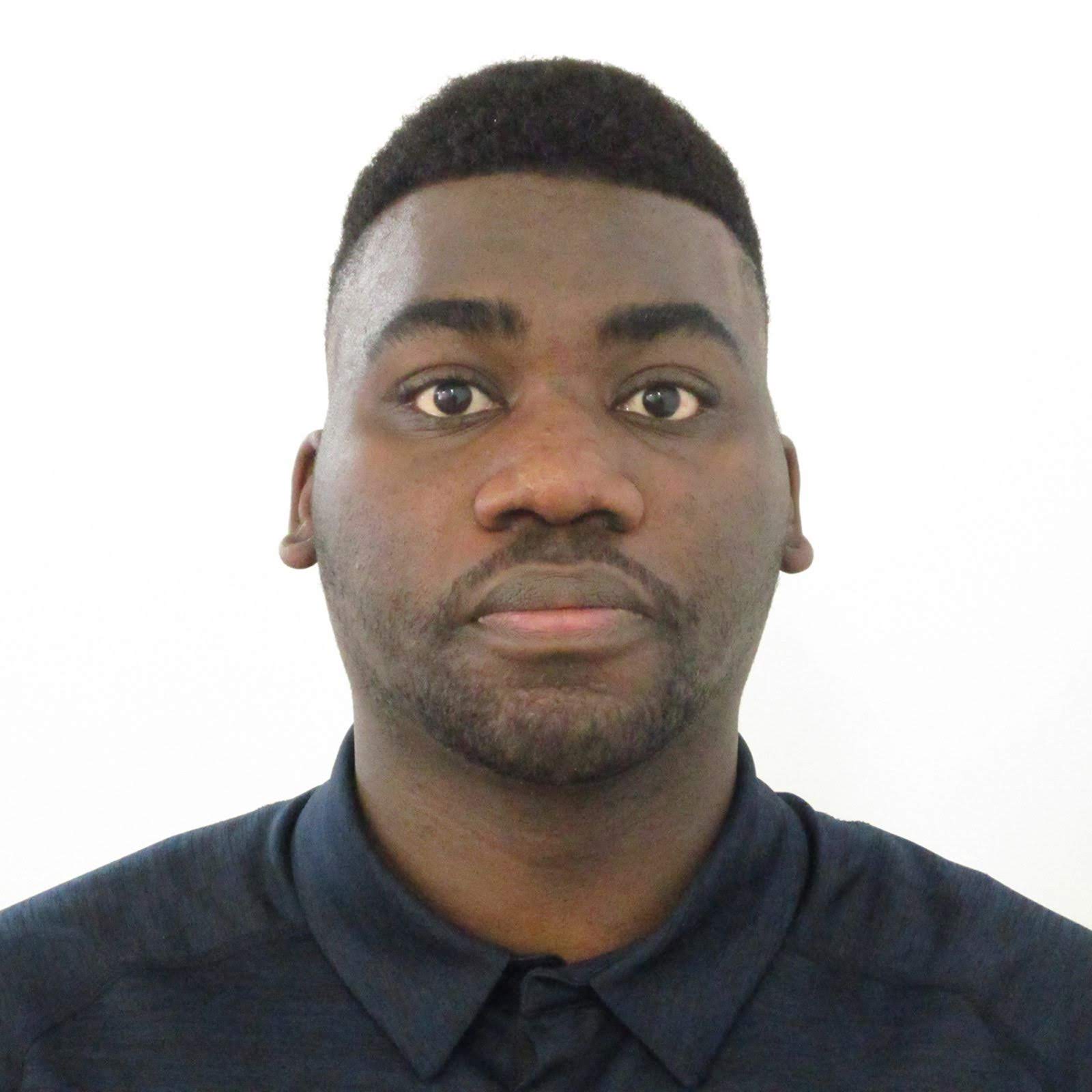}}]{Akanimoh Udombeh} obtained his B.S. in mechanical engineering from Purdue University in 2021. He is currently working towards his M.S. degree in mechanical engineering at The University of Texas at Austin. His interests include mechatronic systems and motion control of dynamic systems.
\end{IEEEbiography}

\begin{IEEEbiography}[{\includegraphics[width=1in,height=1.25in,clip,keepaspectratio]{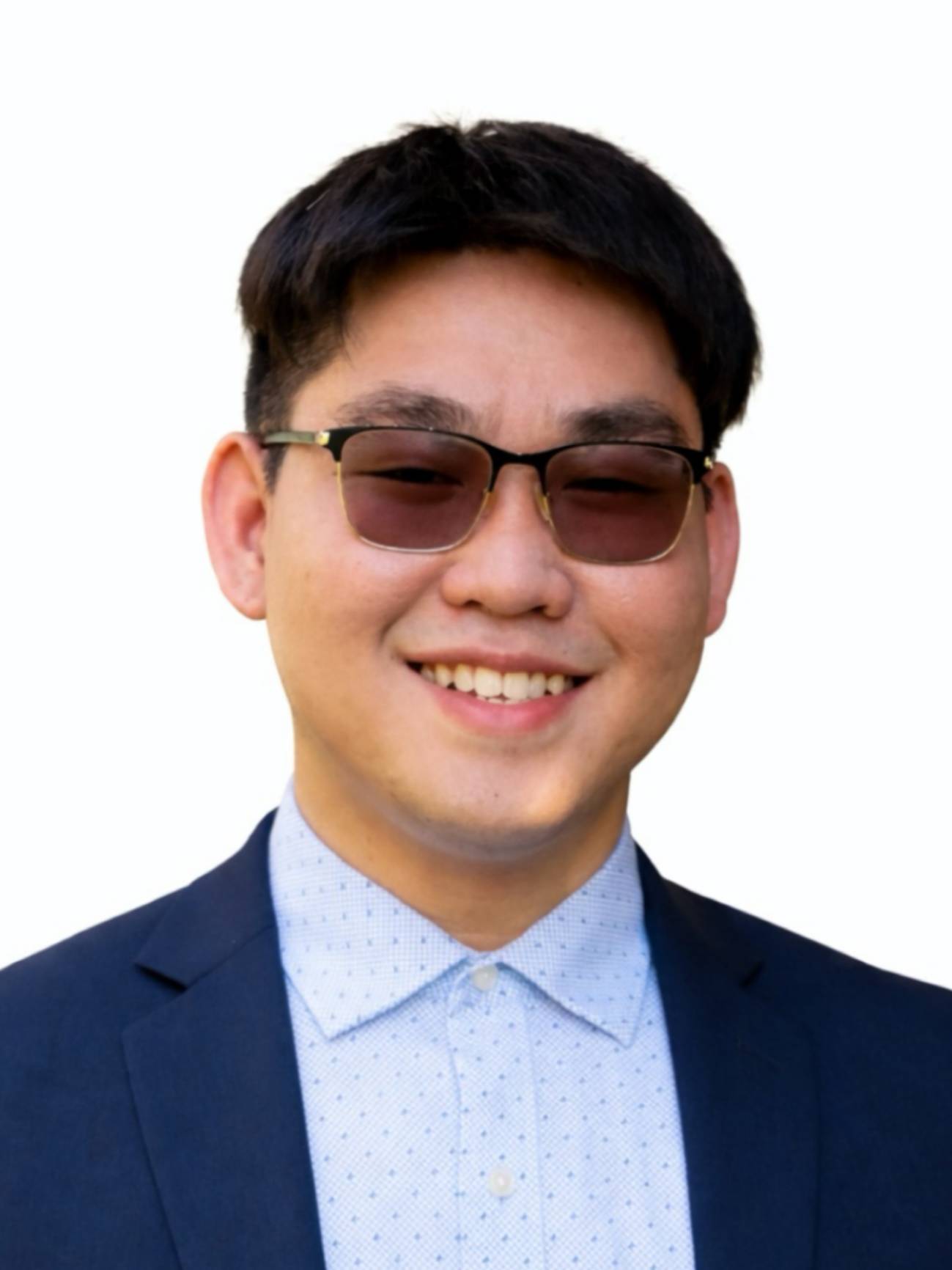}}]{Harrison Teng} is currently pursuing his B.S degree in Mechanical Engineering at the University of Texas at Austin. His academic interests focus on manufacturing, mechanism design, and robotics.
\end{IEEEbiography}

\begin{IEEEbiography}[{\includegraphics[width=1in,height=1.25in,clip,keepaspectratio]{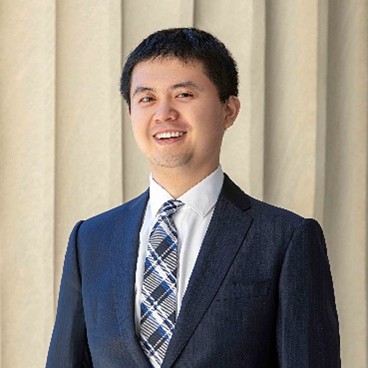}}]{Fangzhou Xia} (Member, IEEE, ASME) received the dual bachelor’s degree in mechanical engineering (ME) from the University of Michigan, Ann Arbor, MI, USA, and in electrical and computer engineering from Shanghai Jiao Tong University, Shanghai, China, in 2015. He received his S.M. in 2017 and Ph.D. in 2020 both in ME from the Massachusetts Institute of Technology (MIT), Cambridge, MA, USA and conduct both of his Postdoc and research scientist training at MIT. 

He joined the Walker Department of Mechanical Engineering at the University of Texas at Austin as an assistant professor in 2024 and directs the MINIMAX Lab. His research interests include precision mechatronics, physical/ computational intelligence, controls, nanorobotics, and instrumentation with applications in biomedical devices, industrial automation and quantum material studies.
\end{IEEEbiography}

\end{document}